\def\year{2019}\relax
\documentclass[letterpaper]{article} 
\usepackage{aaai19}  
\usepackage{times}  
\usepackage{helvet}  
\usepackage{courier}  
\usepackage{url}  
\usepackage{graphicx}  
\usepackage{bm}
\usepackage{amsmath}
\usepackage{amsfonts}
\usepackage{booktabs}
\usepackage{algorithm, algpseudocode}
\usepackage{color}

\graphicspath{{./figures/}}

\newcommand{\1}{\mbox{1}\hspace{-0.25em}\mbox{l}}

\begin{document}
%
\title{Estimation of Dimensions Contributing to Detected Anomalies with Variational Autoencoders}
\author{Yasuhiro Ikeda\textsuperscript{1}, Kengo Tajiri\textsuperscript{1}, Yuusuke Nakano\textsuperscript{1}, Keishiro Watanabe\textsuperscript{1}, Keisuke Ishibashi\textsuperscript{2}\\
\textsuperscript{1}NTT Network Technology Laboratories, NTT Corporation, Tokyo, Japan\\
\textsuperscript{2}Faculty of Natural Sciences, International Christian University, Tokyo, Japan\\
\{yasuhiro.ikeda.sm, kengo.tajiri.bk, yuusuke.nakano.dn, keishiro.watanabe.ry\}@hco.ntt.co.jp, ikeisuke@icu.ac.jp
}
\maketitle
\begin{abstract}
Anomaly detection using dimensionality reduction has been an essential technique for monitoring multidimensional data. Although deep learning-based methods have been well studied for their remarkable detection performance, their interpretability is still a problem. In this paper, we propose a novel algorithm for estimating the dimensions contributing to the detected anomalies by using variational autoencoders (VAEs). Our algorithm is based on an approximative probabilistic model that considers the existence of anomalies in the data, and by maximizing the log-likelihood, we estimate which dimensions contribute to determining data as an anomaly. The experiments results with benchmark datasets show that our algorithm extracts the contributing dimensions more accurately than baseline methods.
\end{abstract}

\section{Introduction}
Anomaly detection in multidimensional data has significant applications in a range of domains such as healthcare monitoring, fraud detection, and system performance monitoring.
Although a number of methods have still been proposed for detecting anomalies from multidimensional data~\cite{zho2018,zon2018},
dimensionality reduction is one of the most important techniques for dealing with high-dimensional data.
Such techniques are based on the assumption that multidimensional data can be embedded into a low dimensional subspace and normal and anomalous data are separable in the embedding~\cite{cha2009}.
There are the following advantages of using the dimensionality reduction for anomaly detection. (i) The anomaly detection can be executed in a semi-supervised fashion by only using normal data and does not require labeled data in anomalous situations, which is generally hard to obtain. (ii) The anomaly score of multidimensional data can be expressed as only one metric: the distance from the normal data subspace. (iii) The anomalies that appear as the collapse of normal relationships among the dimensions can be detected, whereas the surveillance of the individual dimensions will overlook them.
(iv) The ``curse of dimensionality'' in high-dimensional problems will be eased.

Although dimensionality reduction with principal components analysis (PCA) has been widely used for anomaly detection, by virtue of recent rapid advances in deep learning, dimensionality reduction using deep learning such as an autoencoder (AE) has attracted much attention for anomaly detections due to its higher accuracy with non-linear expression. Especially, the variational autoencoder (VAE)~\cite{kin2013} has shown a great potential for obtaining useful latent representations by dimensionality reduction and also for anomaly detection~\cite{xu2018}.

The improvements in detection accuracy with the deep learning-based anomaly detection techniques have been well discussed. However, the interpretability of the techniques is still a problem. The interpretation, that is to say, \textit{why the data is detected as an anomaly}, is of fundamental interest for practical use. Suppose, for example, we monitor medical telemetry data from a patient such as body temperature, heartbeat, and blood pressure data for detecting changes in his/her condition in a healthcare service. A natural approach for applying the deep learning-based anomaly detection is to generate numeric vectors by assuming each monitored data as the value of each dimension in the vectors. Then, if an anomaly is detected, our next question is ``What is happening to the patient?'' To answer that, at least the dimensions contributing to the detected anomaly should be identified. In the deep learning-based anomaly detections, however, the dimensions contributing to the detected anomaly are not directly indicated. Although one might think that the reconstruction error, that is the difference between the input and output data in the multilayer perceptrons (MLPs) of AEs or VAEs, for each dimension can be assumed as the contribution degree to the detected anomalies, the anomalies can affect the reconstruction errors in all dimensions due to their fully connected nature and therefore will cause misestimations. In some prior work~\cite{ike2018,xu2018}, the researchers proposed the algorithms for giving interpretations to deep learning-based anomaly detection techniques. However, the success of their algorithms requires the assumption of the sparsity of the contributing dimensions in all dimensions. Such an assumption limits the application to practical problems in which 
the cause of the anomaly can affect many of the dimensions.

For accurate and interpretable anomaly detections using deep learning, we propose a novel algorithm for estimating the dimensions contributing to the detected anomalies. In our algorithm, VAEs are adopted for an anomaly detection algorithm due to their higher detection accuracy and probabilistic modeling, which is considered to be desirable for giving interpretations. For estimating contributing dimensions, we introduce an approximative probabilistic model based on the trained VAE for exploring a \textit{true latent distribution}, which gives true colors of the detected anomalous data in which the values of the contributing dimensions are fixed to plausible values as normal. We estimate the contributing dimensions via log-likelihood maximization of the model. Through the experiments, we show that the proposed algorithm estimates the contributing dimensions more accurately than conventional approaches.

\section{Related work}
Herein, we mainly review existing work on anomaly detection
based on dimensionality reduction and its interpretation.

PCA has been widely exploited for anomaly detection~\cite{fuj2005,rin2007}. For the interpretation of anomalies, \cite{xu2009} proposed a combination algorithm of PCA and decision trees for visualizing how the anomalies are detected. \cite{jia2012} proposed an algorithm with sparse PCA that computes how much each dimension of multidimensional data contributes to the anomalies. These algorithms, however, assume PCAs for the dimensionality reduction algorithm, and its linear nature limits detection accuracy.

Deep learning-based dimensionality reduction is therefore attracting attention for anomaly detections since it enables non-linear relationships among data to be learned and anomalies to be detected as deviations from them. \cite{sak2014} adopted AEs for anomaly detection with spacecrafts' telemetry data and showed that they achieved better accuracy than linear PCAs. \cite{zho2017} proposed a robust deep AE that eliminates noises from training data and had higher detection accuracy than the isolation forest algorithm. The VAE proposed by~\cite{kin2013} has been widely investigated for obtaining informative latent representation such as text~\cite{xu2017,li2017} and images~\cite{ara2018} and also for anomaly detections. \cite{an2015} evaluated the anomaly detection accuracy using a VAE with the benchmark datasets and showed it had higher accuracy than normal AEs and PCAs. \cite{suh2016} introduced an enhanced VAE for multidimensional time series data to take the temporal dependencies in data into account and demonstrated its superior accuracy to conventional algorithms for time-series monitoring. These studies well investigated the application of deep learning-based anomaly detection algorithms in terms of detection accuracy. However, their interpretations, in other words, the reasons the data are detected as anomalies, were not discussed.

Some prior studies have investigated giving interpretations with deep learning-based anomaly detection algorithms. \cite{ike2018} proposed an algorithm to identify contributing dimensions in anomaly detection with AEs. The algorithm explores a sparse vector of the contribution degree in the input data space under the assumption that the contributing dimensions are fewer in number than all dimensions and fixing their values to plausible values will decrease the anomaly score. Their algorithm, however, assumes that the data is monitored from large systems and the contributing dimensions are a small part of all dimensions. This sparseness assumption is too strong for general problem. \cite{xu2018} enhanced the VAE for monitoring seasonal key performance indicators of web services.
They attempt not only to improve detection accuracy but also to give a theoretical interpretation to the reconstruction error of the VAE by exploiting dependencies in time window. However, the success of the algorithm is also limited to when there are few contributing dimensions. In addition, it is only adoptable for time series data. \cite{sch2017} proposed a novel approach that uses generative adversarial networks (GANs). Their algorithm detects anomalous images in accordance with the difference between a test image and the closest image produced by the generative network and also localizes the anomalous regions. However, the proposed algorithm requires computational cost since the search of the closest image is executed for each detection. Additionally, the algorithm might be difficult to use for general anomaly detection since a GAN requires several tricks for training, as discussed in~\cite{chi2016},
and the efficiency of their algorithm is confirmed only for image detections.

\cite{kuo2016} approached to the interpretation problem of anomalies
in a different way from the above-mentioned model-based methods.
Their framework does not focus on detection of anomalies
but attempts to provide interpretation to given anomalies
using constraint programing with normal data.
Although their framework can be utilized regardless the method of anomaly detection, its combinatorial nature limits the scalability.

In contrast to the prior work discussed above, our proposed algorithm aims at giving interpretation to the anomaly detection with vanilla VAE, which has relatively low computational cost with high dimensional data, for not only seasonal data but also general multidimensional data.
Note that our algorithm is considered to be applicable to other VAE-like anomaly detection techniques as discussed earlier in principle
if the anomaly score is determined in accordance with the likelihood of the test data given the latent variable.

\section{Variational Autoencoder for Anomaly Detection}
The VAE is a generative model with an observed variable $\bm{x}$ and a latent variable $\bm{z}$. Using the model parameters $\bm{\theta}$, we aim to maximize the marginal log-likelihood $\log p_{\bm{\theta}} (\bm{x}) = \log \int p_{\bm{\theta}}( \bm{x} | \bm{z} ) p_{\bm{\theta}}( \bm{z}) d\bm{z}$. Since it is intractable, the VAE introduces a recognition model $q_{\bm \phi} (\bm{z} | \bm {x} )$, which approximates the true posterior $p_{\bm{\theta}} (\bm{z}|\bm{x})$, and the following evidence lower bound (ELBO) is maximized instead;
\begin{equation}\label{eq:vae_elbo}
\begin{split}
\log p_{\bm{\theta}} (\bm{x})  &= \log \int p_{\bm{\theta}} (\bm{x} | \bm{z}) p_{\bm{\theta}} (\bm{z}) d \bm{z} \\
                                 &\geq \int q_{\bm \phi}(\bm{z}|\bm{x}) \log \frac{p_{\bm{\theta}} (\bm{x} | \bm{z}) p_{\bm{\theta}} (\bm{z})}{q_{\bm \phi}(\bm{z}|\bm{x})} d \bm{z}  \\
                                 &= \mathbb{E}_{q_{\bm \phi} (\bm{z}|\bm{x})} [ \log p_{\bm{\theta}} (\bm{x} | \bm{z} ) ] - D_{KL} [ q_{\bm \phi}(\bm{z}|\bm{x}) || p_{\bm{\theta}} (\bm{z}) ].
\end{split}
\end{equation}
By using MLPs for constructing $q_{\bm \phi} (\bm{z}|\bm{x})$ as an encoder and $p_{\bm{\theta}} (\bm{x}|\bm{z})$ as a decoder, and by applying the reparametarization trick~\cite{kin2013}, $q_{\bm \phi} (\bm{z}|\bm{x})$, and $p_{\bm{\theta}} (\bm{x}|\bm{z})$ are estimated simultaneously by using backpropagation. Encoder $q_{\bm \phi} (\bm{z}|\bm{x})$ is often assumed as a multivariate Gaussian with a diagonal covariance, and the output of the MLP corresponds to its mean and standard deviation. Decoder $p_{\bm{\theta}} (\bm{x}|\bm{z})$ is generally assumed as a multivariate Gaussian with a diagonal covariance or Bernoulli, and the output of the MLP corresponds to their parameters.

For using the VAE as an anomaly detection algorithm, the VAE is trained by using only normal data, and anomaly scores of test data are expressed by the negative of the ELBO since the closer the test data is to the trained distribution $p_{\bm{\theta}} (\bm x)$ with normal data, the higher the ELBO tends to be and vice versa. The VAE can also be assumed as a regularized AE since the second term in Eq.~(\ref{eq:vae_elbo}) acts as a regularization term to the encoder.
It prevents overfitting, and the VAE is expected to have higher detection accuracy than vanilla AEs. However, as also discussed by~\cite{xu2018}, although the anomaly score is computed in accordance with the ELBO calculated by the test data and Eq.~(\ref{eq:vae_elbo}), it does not necessarily express the correct ELBO of the log-likelihood of the test data since the VAE is trained with Eq.~(\ref{eq:vae_elbo}) under the assumption that all the data is normal and therefore the result computed with the anomalous data is biased. This is further discussed in the next section.

\section{Proposed Algorithm}
Suppose that test data $\bm{x}'= \{ x_1',...,x_N' \}$ is detected as an anomaly by the VAE. Then, our objective is to identify the dimensions contributing to the anomaly from $i=1,...,N$. Since the ELBO of the VAE includes the expectation of the log-likelihood of each dimension as $\mathbb{E}_{q_{\bm \phi}(\bm z| \bm x)} [\log p_{\bm{\theta}} (\bm x | \bm z) ] = \sum_{i=1}^N \mathbb{E}_{q_{\bm \phi}(\bm z| \bm x)} [ \log p_{\bm{\theta}} ( x_i | \bm z ) ]$, the negative of which is also called reconstruction error, one might think that the contribution of each dimension to the detected anomaly can be directly derived from the reconstruction error of each dimension $-\mathbb{E}_{q_{\bm \phi}(\bm z| \bm x)} [ \log p_{\bm{\theta}} ( x_i | \bm z ) ]$. However, since the VAE is trained by only normal data, and $q_{\bm \phi}(\bm z | \bm x)$ is a recognition model that infers the posterior $p_{\bm{\theta}} (\bm z | \bm x)$ under the assumption that $\bm x$ is normal data,
$q_{\bm \phi}(\bm z| \bm x)$ calculated by using the anomalous data $\bm{x}'$ includes errors.
Estimating the contributing dimensions by using the log-likelihood with the deviating latent distribution will cause false positives (FPs) and false negatives (FNs), which are numerically examined in the experiments.
Thus we define the true latent distribution $\hat{q}(\bm{z})$ that expresses a latent representation of data $\hat{\bm{x}}$, where $\hat{\bm{x}}$ is how the anomalous data $\bm{x}'$ would look if it were normal. Since $\hat{q}(\bm{z})$ and $\hat{\bm{x}}$ are not known directly, we aim to approximate $\hat{q}(\bm{z})$ with the trained VAE and the anomalous test data.

Our algorithm is based on the following assumption.

\noindent \textbf{Assumption.} Suppose that $\hat{\bm{x}}$ is a normal data made by amending the values of contributing dimensions of the anomalous data $\bm{x}'$ to plausible values in a normal situation and $\hat{q}(\bm z)$ is a latent distribution that maximizes the ELBO of $\log p_{\bm \theta}(\hat{\bm{x}})$ with the trained decoder. Then, a set of contributing dimensions $\Psi$ is extracted by a threshold value $\beta$ as $\Psi = \{ i | \mathbb{E}_{\hat{q}(\bm z)} [ p_{\bm{\theta}} ( x'_i | \bm z ) ] < \beta \}$ and therefore;
\begin{equation}
\begin{split}
x'_i = \hat{x}_i \quad &\mathrm{If} \, i \notin \Psi \\
x'_i \neq \hat{x}_i \quad &\mathrm{Else}.
\end{split}
\end{equation}

The assumption expresses that all the expected probabilities $\mathbb{E}_{\hat{q}(\bm z)} [ p_{\bm{\theta}} ( x'_i | \bm z ) ]$ of the contributing dimensions computed by the true latent distribution are smaller than those of the dimensions irrelevant to the anomaly. According to this assumption, we try to estimate $\Psi$ by exploring true latent distribution $\hat{q}(\bm z)$. Although $\hat{q}(\bm z)$ can be obtained as a latent distribution that maximizes the ELBO of $\log p_{\bm \theta}(\hat{\bm{x}})$, $\hat{\bm{x}}$ is unknown. Therefore, we approximately explore $\hat{q}(\bm z)$ with observed anomalous data $\bm{x}'$ and the trained VAE as follows. From the assumption, the conditional probability $p_{\bm{\theta}} (\hat{\bm{x}} | \bm z )$ can be factorized as;
\begin{equation} 
p_{\bm{\theta}} (\hat{\bm{x}} | \bm z ) = \prod_{i \in \Psi} p_{\bm{\theta}} (\hat{x}_i|\bm z) \prod_{i \notin \Psi} p_{\bm{\theta}} (x'_i|\bm z).
\end{equation}
Since $\hat{\bm{x}}$ is not known, we approximate $p_{\bm{\theta}} (\hat{x}_i|\bm z) \; \forall i \in \Psi$ by the geometric mean of the conditional probability of the other irrelevant normal dimensions as $\bar{p} = [ \prod_{i \notin \Psi} p_{\bm{\theta}} (x'_i|\bm z) ]^{\frac{1}{N-|\Psi|}}$.
Then, the marginal log-likelihood of $\log p_{\bm{\theta}}(\hat{\bm x})$ and its ELBO with latent distribution $q (\bm z)$ can be approximated as follows;

\begin{equation} \label{eq:mll}
\begin{split}
\log p_{\bm{\theta}} (\hat{\bm x}) =& \log \int p_{\bm{\theta}} ( \hat{\bm x} | \bm{z} ) p_{\bm{\theta}} (\bm{z} ) d\bm{z} \\
                            \approx& \log \int \Bigl[ \Pi_{i \notin \Psi} p_{\bm{\theta}} (x_i' | \bm{z} ) \Bigr]^{\frac{N}{N-|\Psi|}} p_{\bm{\theta}} (\bm{z} ) d\bm{z}. \\
                            \geq& \frac{N}{N-|\Psi|}  \mathbb{E}_{q(\bm{z})} \Biggl[ \sum_{i \notin \Psi} \log  p_{\bm{\theta}} (x'_i | \bm{z}) \Biggr] \\
					&- D_{KL} [ q(\bm{z}) || p_{\bm{\theta}} (\bm{z}) ].
\end{split}
\end{equation}
From the assumption, $q (\bm{z})$ that maximizes the ELBO of Eq.(\ref{eq:mll}), which we denote as $\dot{q} (\bm{z})$, can be assumed as the approximative latent distribution of $\hat{q} (\bm z)$. The contributing dimensions are thus estimated in accordance with the expected probability $\mathbb{E}_{\dot{q} (\bm z )} [ p_{\bm{\theta}} (x_i' | \bm{z}) ]$. The ELBO, however, includes $\Psi$ determined by $\hat{q} (\bm z)$ and is not known a priori. Therefore, during the exploration of $q (\bm{z})$, we further approximate $\Psi$ as the set of least $K$ dimensions in $\mathbb{E}_{q(\bm z )} [ p_{\bm{\theta}} (x_i' | \bm{z}) ]$, denoted as $\Phi$, where $K$ is heuristically determined as explained later. After we obtain $\dot{q} (\bm{z})$, we finally estimate the contributing dimensions as $\Psi = \{ i | \mathbb{E}_{ \dot{q} (\bm z )} [ p_{\bm{\theta}} (x'_i | \bm{z}) ] < \beta \}$. Note that we do not simply assume $\Phi = \{ i | \mathbb{E}_{ q (\bm z )} [ p_{\bm{\theta}} (x'_i | \bm{z}) ] < \beta \}$ during the exploration of $q (\bm z)$ since it drastically fluctuates the objective function and will cause trapping to local maxima. Consequently, the objective function for maximization becomes as follows:
\begin{equation} \label{eq:elbo}
\max_{q(\bm z)} \frac{N}{N-K}  \mathbb{E}_{q(\bm{z})} \Biggl[ \sum_{i \notin \Phi} \log  p_{\bm{\theta}} (x'_i | \bm{z}) \Biggr] - D_{KL} [ q(\bm{z}) || p_{\bm{\theta}} (\bm{z}) ].
\end{equation}

\begin{algorithm}[t]
\caption{Pseudo code of the proposed algorithm}
\label{alg:pro}
\begin{algorithmic}[1]
\State Set initial $\bm{\mu}_{\bm{z}}, \bm{\sigma}_{\bm{z}}$ with trained encoder $q_{\bm \phi} (\bm{z}|\bm{x}')$
\State Set initial $K = \sum_{i=1}^N \1 (\mathbb{E}_{q_{\bm \phi}(\bm{z}|\bm{x}')} [ p_{\bm{\theta}} (x_i' | \bm{z} ) ] < \beta ] )$
\While{maximizing Eq.(\ref{eq:elbo}) converges}
\State Update $\Phi$ in accordance with $\bm{\mu}_{\bm{z}}$ and $\bm{\sigma}_{\bm{z}}$
\State Update $\bm{\mu}_{\bm{z}}, \bm{\sigma}_{\bm{z}}$ to increase Eq.(\ref{eq:elbo})
\If{Eq.(\ref{eq:elbo}) converged less than $\gamma$}
\State Update $K = K + K_{inc}$
\State Reset $\bm{\mu}_{\bm{z}}, \bm{\sigma}_{\bm{z}}$ with trained encoder $q_{\bm \phi} (\bm{z}|\bm{x}')$
\EndIf
\EndWhile
\State Estimate contributing dimensions as
\Statex $\Psi = \{ i | \mathbb{E}_{\mathcal{N}(\bm{z}; \bm{\mu}_{\bm z}, \bm{\sigma}_{\bm z}^2 \bm{I})} [ p_{\bm{\theta}} (x_i' | \bm{z} ) ] < \beta \}$
\State Output $\Psi$
\end{algorithmic}
\end{algorithm}

The pseudo code of the proposed algorithm is depicted in Algorithm~\ref{alg:pro}. Similar to the general VAEs, we assume $q(\bm z)$ as multivariate Gaussian with a diagonal covariance $\mathcal{N}(\bm{z}; \bm{\mu}_{\bm z}, \bm{\sigma}_{\bm z}^2 \bm{I})$. Thus, the objective is to obtain $\bm{\mu}_{\bm z}, \bm{\sigma}_{\bm z}$ that maximizes Eq.(\ref{eq:elbo}), and its initial values are determined by the trained encoder of the VAE. Note that estimating contributing dimensions by using initial $q(\bm z)$ corresponds to the estimation with the reconstruction error of the vanilla VAE. We first determine $K$ in accordance with the reconstruction error, and if Eq.(\ref{eq:elbo}) converges under the threshold, it is incremented and then the maximization is executed again. We set $\gamma$ as the mean value of the ELBO with the trained VAE and the training data. In this paper, $\bm{\mu}_{\bm z}, \bm{\sigma}_{\bm z}$ is updated by a simple gradient descent method.

Not only estimating the contributing dimensions, we can also calculate the \textit{contribution degree} that expresses how much a dimension contributes to the detected anomaly. Since the MLP of decoder $p_{\bm{\theta}} (\bm x | \bm z )$ outputs the parameters of the assumed distribution, we can calculate how much the contributing dimensions deviate from the normal situation as the deviation from the distribution with the expected parameters. For example, if the decoder is assumed as a multivariate Gaussian, the contribution degree of $x'_i$ can be calculated as $\frac{x'_i - \bar{\mu}_i}{\bar{\sigma}_i}$, where $\bar{\mu}_i$ and $\bar{\sigma}_i$ are the expectations of the mean and standard deviation values calculated by the MLP of the decoder over the obtained distribution $\mathcal{N}(\bm{z}; \bm{\mu}_{\bm z}, \bm{\sigma}_{\bm z}^2 \bm{I})$.

Since our proposed algorithm is just an approximate estimation, two concerns arise. (i). Due to the assumption and several approximations, $q(\bm{z})$ that maximizes Eq.~(\ref{eq:elbo}) does not necessarily correspond to $\hat{q}(\bm z)$ and therefore produce FPs and FNs. (ii). Even if the optimal $q(\bm{z})$ corresponds to $\hat{q}(\bm z)$, the optimization may be trapped to local maxima due to the discontinuity and multi-modality of the objective function (Eq.~(\ref{eq:elbo})). The algorithm, however, shows better estimation accuracy than conventional methods as discussed in the experiments. We consider why the algorithm works well with respect to the above concerns. (i). The situation in which optimizing Eq.~(\ref{eq:elbo}) produces FPs and FNs can be divided into two cases. One is that the dimensions whose log-likelihoods are maximized include anomaly dimensions. This can be caused when the values of the contributing dimensions are more plausibly considered as normal. Such anomalies, however, are unlikely to occur often. The other is that although the maximized dimensions did not include any contributing dimensions, the obtained latent distribution deviates from $\hat{q}(\bm z)$. This is likely to occur when the size of the ignored dimensions set $\Phi$, that is $K$ in the algorithm, is excessively larger than the number of truly contributing dimensions. Therefore, $K$ should be set to a plausible value and carefully updated in Algorithm~\ref{alg:pro}. Although we set $K_{inc}$ as 10\% of the number of the whole dimensions, its adequate determination considering both the estimation accuracy and the computational cost should be discussed in future work. Note that the latter case is of course unavoidable when the contributing dimensions account for most of the dimensions. (ii). Even though we may obtain sub-optimal $q(\bm z)$ depending on the initial parameters in exploration, as discussed in several related studies such as~\cite{zha2018}, VAE-like autoencoders that regularize the representation in the latent space have a property that projects similarly-constructed data to nearby in the latent space. According to the property, the initial parameters given by encoding the anomaly data can be close to the optimal solution
if the ratio of contributing dimensions is not too large.

Note that exploring a true latent distribution is a similar approach to the GAN-based image inpainting algorithm by~\cite{yeh2017}. The algorithm, which inpaints the hole area of a corrupted image, explores the latent variable so that the distance between the image produced with the generative model and the corrupted image other than the hole area becomes closest. Our algorithm also explores the latent distribution so that maximizing the log-likelihood of the conditional probability while ignoring the contributing dimensions to the anomaly. However, since the contributing dimensions themselves are the objective to estimate and are not known in advance, we approximate them and update for each iteration in the exploration. The anomalous image detection algorithm by~\cite{sch2017} discussed in the related work is also inspired by the image inpainting algorithm. Although the algorithm just explores a latent variable that generates the closest image to the test image and does not consider the existence of anomalous areas, they leverage the discriminator so that the explored latent variable fits to the learned distribution with normal images and therefore the exploration is not biased to the anomalous area. However, as already discussed, the algorithm becomes complex and time consuming.

\begin{table*}[t]
\caption{Parameters of VAE/AE model. We used Chainer~\cite{tok2015} as a deep learning framework.}
\small
\begin{tabular}{c|c|c}
\toprule[0.5mm]
Parameters & VAE & AE \\
\midrule
\# of layers & 5 (2 for encoder and decoder, 1 for latent space.) & 5 \\
\midrule
ratio of size to input in each layer & 0.7,0.5,0.1,0.5,0.7 & 0.7,0.5,0.1,0.5,0.7 \\
\midrule
activation functions in each layer & tanh, tanh, identity, tanh, tanh & ReLU, ReLU, identity, ReLU, ReLU \\
\midrule
\# of epochs & \multicolumn{2}{|c}{100}  \\
\midrule
batch size & \multicolumn{2}{|c}{16 (Arrhythmia), 64 (MNIST and Musk), and 128 (NSL-KDD)}  \\
\midrule
weight decay & \multicolumn{2}{|c}{1E-3} \\
\midrule
dropout ratio & \multicolumn{2}{|c}{0.2 in input layer and 0.5 in hidden layers (adopted only for benchmarks other than NSL-KDD)} \\
\bottomrule[0.5mm]
\end{tabular}
\label{tab:par}
\end{table*}

\section{Experiments}\label{sec:exp}
We evaluate our proposed algorithm with four benchmark datasets. In the first experiment, we evaluate how accurately it estimates the contributing dimensions with artificially introduced anomalies. After that, we also apply our algorithm to the labeled anomaly data and discuss if the estimated contributing dimensions are plausible.
Note that we leave the performance evaluation of anomaly detection to the appendix since it is not the main scope of this paper.
For evaluation, we use following benchmark datasets.
\begin{description}
\item[Arrhythmia.] The Arrhythmia dataset~\cite{guv1997} obtained from the UCI repository\footnote[1]{https://archive.ics.uci.edu/ml/index.php} is composed of multiple medical data including cardiac rhythms, and the objective is to detect the presence of cardiac arrhythmia. Each data belongs to one normal class or one of 15 anomaly classes. Since features of several attributes are missing in the original data, we use the preprocessed dataset for anomaly detection in ODDS repository\footnote[2]{http://odds.cs.stonybrook.edu/} in which the missing and categorical features are discarded and the 16 classes are aggregated into two classes: normal and outlier (defined as anomaly in this paper). Consequently, the dataset is composed of 386 normal and 66 anomalous data with 274 dimensions.
\item[MNIST.] The MNIST dataset~\cite{lec1998} is a representative benchmark of handwritten digits. We also use a preprocessed datafrom ODDS repository, where the dataset is converted so that the digit-zero class becomes a normal class and the sampled digit-six class becomes anomalies. In addition, the number of dimensions is downsized to 100 by randomly selecting from 700 original features. Consequently, the dataset is composed of 6,903 normal and 700 anomalous data.
\item[Musk.] The Musk dataset~\cite{agg2015} describes a set of molecules, and the objective is to detect musks from non-musks. We also use preprocessed data from ODDS repository and consequently the dataset is composed of 2,965 normal and 97 anomalous data with 166 dimensions.
\item[NSL-KDD.] The NSL-KDD dataset~\cite{tav2009} is the amended version of the KDD Cup 1999 dataset\footnote[3]{http://kdd.ics.uci.edu/databases/kddcup99/kddcup99.html} the task of which is to detect attack connections by using several network connection features. We use 20$\%$ subset data that consists 13,448 normal and 5,813 anomalous data. Since it contains 38 continuous and 3 symbolic features, we convert the symbolic features into one-hot vectors and consequently obtained the dataset with 118 dimensions.
\end{description}
The datasets are preprocessed so that the values of each dimension are standardized in accordance with the mean and standard deviation values of the dimension in the training data. Therefore, we adopted a multivariate Gaussian for the decoder of the VAE. Although our algorithm also works with a Bernoulli decoder in principle, we will investigate that in future work. The parameters of the anomaly detection algorithms with the AE and the VAE are described in Table~\ref{tab:par}. Note that the dropout is not applied to the model in the evaluation with the NSL-KDD dataset since it strongly degrades the detection accuracy. The reason can be considered the sparseness of the input data due to the one-hot vector representation. For the proposed estimation algorithm of contributing dimensions, we set $\beta$ as the first percentile of the expected probability $\mathbb{E}_{q_{\bm \phi} (\bm z | \bm x )} [ p_{\bm{\theta}} (x_i | \bm{z}) ]$ calculated with the trained VAE and the training data.

\begin{table*}[t]
\caption{Estimation accuracy. Percentages on top denote the ratio of $M$ to the number of dimensions. Note that the ratio with NSL-KDD is up to 30\% since $M$ dimensions are selected only from 38 continuous dimensions.}
\small
\begin{center}
\begin{tabular}{c|c|cccccc|cccccc}
\toprule[0.5mm]
 &  & \multicolumn{6}{c|}{Arrhythmia} & \multicolumn{6}{c}{MNIST}  \\
Method & Metric & 10\% & 20\% & 30\% & 40\% & 50\% & 60\% & 10\% & 20\% & 30\% & 40\% & 50\% & 60\% \\
\midrule
 & \# of FPs & 6.35 & 20.15 & 8.1 & 9 & 21.7 & 16.5 & 5.45 & 6.25 & 3.85 & 6.4 & 7.1 & 5.4 \\
$AE_{SO}$ & \# of FNs & \textbf{0.05} & \textbf{0.1} & \textbf{0.1} & \textbf{0.1} & \textbf{0.1} & \textbf{0.1} & \textbf{0.85} & \textbf{2.75} & \textbf{3.45} & \textbf{4.85} & \textbf{5.8} & \textbf{7.05} \\
 & $F_1$ & 0.9 & 0.85 & 0.95 & 0.96 & 0.93 & 0.95 & 0.76 & 0.8 & 0.88 & 0.86 & 0.87 & 0.89 \\
\midrule
 & \# of FPs & 1.3 & 1.5 & 1.45 & 1.3 & \textbf{0.6} & 0.65 & 0.45 & 0.25 & 0.5 & \textbf{0.25} & 0.15 & 0.3 \\
$VAE_{rec.}$ & \# of FNs & 2.7 & 6.65 & 17.45 & 9.25 & 14.3 & 32.15 & 2.55 & 3.9 & 6.65 & 10.25 & 10.55 & 14.7 \\
 & $F_1$ & 0.92 & 0.92 & 0.86 & 0.94 & 0.94 & 0.87 & 0.83 & 0.88 & 0.86 & 0.84 & 0.88 & 0.85 \\
\midrule
 & \# of FPs & \textbf{1.15} & \textbf{0.95} & \textbf{1.05} & \textbf{1.05} & \textbf{0.6} & \textbf{0.5} & \textbf{0.4} & \textbf{0.1} & \textbf{0.25} & 0.3 & \textbf{0.05} & \textbf{0.25} \\
$VAE_{pro.}$ & \# of FNs & 0.85 & 1.3 & 1.65 & 3.4 & 4.85 & 3.9 & 1.9 & 3.4 & 4.65 & 7.35 & 8.45 & 10.2 \\
 & $F_1$ & \textbf{0.96} & \textbf{0.98} & \textbf{0.98} & \textbf{0.98} & \textbf{0.98} & \textbf{0.99} & \textbf{0.87} & \textbf{0.9} & \textbf{0.91} & \textbf{0.89} & \textbf{0.91} & \textbf{0.9} \\
\midrule
 &  & \multicolumn{6}{c|}{Musk}  & \multicolumn{6}{c}{NSL-KDD} \\
&  & 10\% & 20\% & 30\% & 40\% & 50\% & 60\% & 10\% & 20\% & 30\% & 40\% & 50\% & 60\% \\
\midrule
 & \# of FPs & 8.4 & 11.8 & 25.3 & 25.75 & 20.4 & 14 & 12.75 & 19.25 & 21.71 & - & - & - \\
$AE_{SO}$ & \# of FNs & \textbf{0} & \textbf{0.05} & \textbf{0} & \textbf{0} & \textbf{0.2} & \textbf{0.15} & \textbf{0.1} & \textbf{0.55} & \textbf{0.76} & - & - & - \\
 & $F_1$ & 0.82 & 0.85 & 0.82 & 0.84 & 0.89 & 0.93 & 0.69 & 0.71 & 0.75 & - & - & - \\
\midrule
 & \# of FPs & 0.55 & 0.75 & 1.1 & 0.5 & 1.4 & 0.4 & 4.3 & 3 & 1.94 & - & - & - \\
$VAE_{rec.}$ & \# of FNs & 0.65 & 1.3 & 1.85 & 4 & 4.4 & 5.25 & 4.65 & 8.6 & 12.94 & - & - & - \\
 & $F_1$ & 0.96 & 0.97 & 0.97 & \textbf{0.97} & 0.96 & \textbf{0.97} & 0.58 & 0.65 & 0.68 & - & - & - \\
\midrule
 & \# of FPs & \textbf{0.4} & \textbf{0.5} & \textbf{0.4} & \textbf{0.4} & \textbf{0.55} & \textbf{0.35} & \textbf{1.45} & \textbf{0.45} & \textbf{0.88} & - & - & - \\
$VAE_{pro.}$ & \# of FNs & 0.5 & 1.1 & 1.6 & 3.7 & 3.45 & 4.85 & 2.25 & 5.4 & 7.35 & - & - & - \\
 & $F_1$ & \textbf{0.97} & \textbf{0.98} & \textbf{0.98} & \textbf{0.97} & \textbf{0.98} & \textbf{0.97} & \textbf{0.79} & \textbf{0.78} & \textbf{0.81} & - & - & - \\
\bottomrule[0.5mm]
\end{tabular}
\end{center}
\label{tab:acc_cont}
\end{table*}

\subsection{Experiment with synthetic anomaly data}\label{sec:exp_syn}
We evaluate the estimation accuracy of the contributing distributions to the detected anomalies by deep learning-based techniques with synthetic anomalies. In the experiments, we use only normal data in the benchmark dataset and evaluated the estimation accuracy through the following procedure. We first randomly extract 90$\%$ of the normal data as the training data and the rest as the candidate test data. After training the anomaly detection model with the training data, we choose a test data from candidate data that achieves an anomaly score lower than $\gamma$ for generating synthetic anomaly data that purely includes artificial anomaly. Then we insert an anomaly to the chosen test data by selecting $M$ dimensions randomly and by fluctuating the values to $r \sigma_i$, where $r$ is a random variable uniformly distributed between $[3,5]$ and $[-5,-3]$, and $\sigma_i$ is the standard deviation of $i$-th dimension in the training data. Although $M$ dimensions are randomly selected, the candidates are the dimensions in which $\sigma_i$ is a positive value (not a constant in training data) and continuous data in NSL-KDD. The randomly selected dimensions in the synthetic anomaly data are then estimated by our proposed algorithm ($VAE_{pro.}$) and two baseline methods: sparse optimization with an AE by~\cite{ike2018} ($AE_{SO}$), the details of which is described in the appendix, and the reconstruction error with the vanilla VAE ($VAE_{rec.}$). The estimated dimensions with $VAE_{rec.}$ correspond to $K$ dimensions in second line of Algorithm~\ref{alg:pro}. The adopted anomaly detection models are VAEs for $VAE_{pro.}$ and $VAE_{rec.}$, and an AE for $AE_{SO}$. The evaluation metrics are the mean of the number of FPs and FNs and the $F_1$ score through 20 runs. We use the number of FPs and FNs, not precision and recall, since precision and recall directly depend on the number of the targets of the detection and therefore the trend according to the variation of the number of the targets for each model becomes obscure. Although the $F_1$ score also suffers from that, we show it as a comprehensive metric for comparison among the algorithms with the same number of target dimensions.

Table~\ref{tab:acc_cont} shows the comparison results of the estimation accuracies.
Although we also evaluated a naive method which estimates the contributing dimensions in accordance with the deviation from the distribution of the individual dimensions in the training data,
the $F_1$ scores were around 0.6 and therefore we omit the results due to space limitation. 
From the result, we find that the proposed algorithm generally outperforms other baseline methods. Especially, our algorithm can depress the number of FPs, which is particularly high in $AE_{SO}$ with a larger ratio of the contributing dimensions. On the other hand, the number of FNs with the proposed algorithm tends to be higher than that with $AE_{SO}$. This can be explained from the characteristics of the VAE based approach that estimates the contributing dimensions in accordance with the conditional probability $p_{\bm{\theta}} (x_i'|\bm{z})$ and therefore the small deviation can be overlooked, whereas $AE_{SO}$ estimates them in accordance with the absolute value of the distance each dimensions are moved in sparse optimization. Note that, contrary to intuition, the $F_1$ score with $AE_{SO}$ increases as the ratio of fluctuated dimensions increases in spite of its sparseness assumption. The reason is that the estimation with $AE_{SO}$ is sensitive and prone to produce FPs, but the candidate dimensions of FPs decrease as the number of the target dimensions increases. Such a positive effect on precision can be thought to more than compensate for the negative effect caused by the deviation from the sparseness assumption at a point.

\begin{figure}[t]
  \begin{center}
   \includegraphics[width=60mm]{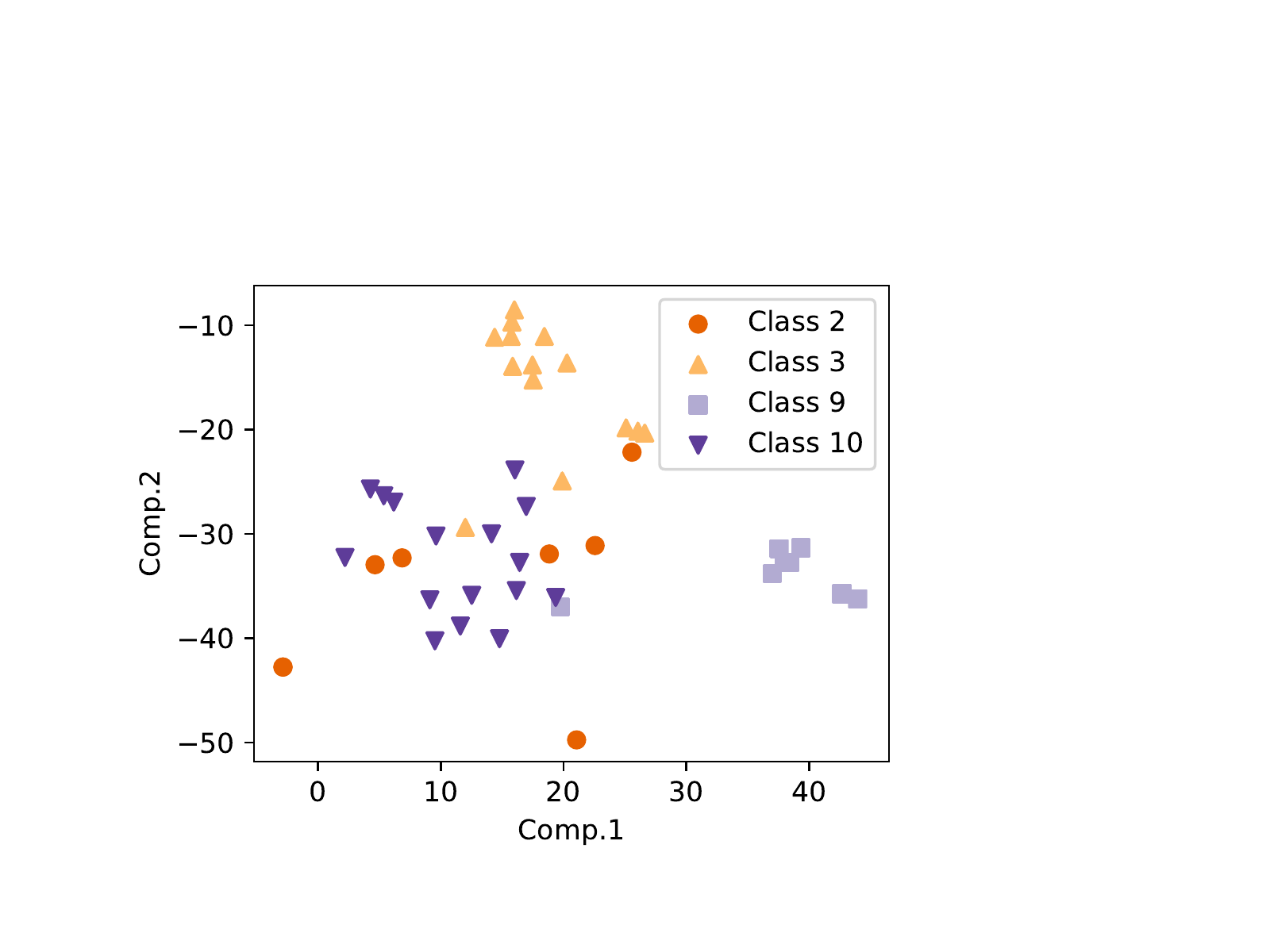}
  \end{center}
  \caption{t-SNE plot of contribution degree.}
  \label{fig:tsne}
\end{figure}

\begin{table*}[h]
\caption{Characteristics of labeled anomalies and estimated contributing dimensions. The contributing dimensions are listed in descending order of their absolute values of median contribution degrees over the detected test data. Up (down) arrow indicates that the median contribution degree was positive (negative). The dimensions that seem to be related to the characteristic of the class are highlighted in bold.}
\begin{center}
\scalebox{0.85}{ 
\begin{tabular}{c|c|c}
\toprule[0.5mm]
Classes & Characteristic\footnote[4] & Estimated contributing dimensions (top 5) \\
\midrule
 \begin{tabular}{c} Class 2 \\ Ischemic changes \\ (Coronary Artery Disease) \end{tabular} & \begin{tabular}{c} \textit{The symptoms of coronary artery disease are} \\ \textit{quite different from person to person which can} \\
\textit{make it a challenge for doctors to determine if a patient's} \\ \textit{symptoms are from coronary blockage (angina) or not.} \end{tabular} & \begin{tabular}{c} $\downarrow$chDI\_TwaveAmp, $\uparrow$chAVR\_TwaveAmp, \\ $\downarrow$chV5\_TwaveAmp, $\downarrow$chV6\_TwaveAmp, \\
       $\downarrow$chDI\_JJwaveAmp \\ (contribution degrees were relatively low) \end{tabular}  \\
\midrule
 \begin{tabular}{c} Class 3 \\ Old Anterior \\ Myocardial Infarction \end{tabular} & \begin{tabular}{c} \textit{ST segment elevation} \\ \textit{in the anterior leads (V3 and V4) at the J point,}  \\ \textit{Reciprocal ST segment depression} \\ \textit{in the inferior leads (II, III and aVF)}, \\ \textit{Abnormalities of the Q waves are} \\ \textit{mostly indicative of myocardial infarction...} \end{tabular} & \begin{tabular}{c} \bf{$\downarrow$chV3\_QwaveAmp}, \bf{$\uparrow$chV3\_Qwave}, \\ \bf{$\downarrow$chV4\_QwaveAmp}, \bf{$\uparrow$chV4\_Qwave}, \\
       \bf{$\uparrow$chV2\_Qwave}\end{tabular} \\
\midrule
 \begin{tabular}{c} Class 9 \\ Left bundle branch block \end{tabular} & \begin{tabular}{c} \textit{QRS duration greater than 120 milliseconds,} \\ \textit{Absence of Q wave in leads I, V5 and V6,} \\ \textit{Monomorphic R wave in I, V5 and V6,} \\ \textit{ST and T wave displacement} \\ \textit{opposite to the major deflection of the QRS complex} \end{tabular} & \begin{tabular}{c} \bf{$\uparrow$chV5\_Rwave}, \\ $\uparrow$chV5\_intrinsicReflecttions, $\downarrow$chV2\_QRSA, \\ \bf{$\uparrow$QRSduration}, \bf{$\uparrow$chV6\_Rwave}\end{tabular} \\
\midrule
 \begin{tabular}{c} Class 10 \\ Right bundle branch block \end{tabular} & \begin{tabular}{c} \textit{QRS duration greater than 120 milliseconds,} \\
\textit{rsR' ``bunny ear'' pattern} \\ \textit{in the anterior precordial leads (leads V1-V3),} \\
\textit{Slurred S waves in leads I, aVL and frequently V5 and V6} \end{tabular} & \begin{tabular}{c} \bf{$\uparrow$chV1\_RPwaveAmp}, \bf{$\uparrow$chV1\_RPwave}, \\ $\uparrow$chV1\_intrinsicReflecttions, \\ $\uparrow$chV1\_QRSA, $\downarrow$chV2\_Swave \end{tabular} \\
\bottomrule[0.5mm]
\end{tabular}
}
\end{center}
\label{tab:cha}
\end{table*}

\subsection{Experiment with labeled anomaly data}\label{sec:exp_lab}
Finally, we qualitatively confirm if our proposed algorithm can extract plausible dimensions as the dimensions contributing to the anomaly with labeled anomaly data. In this experiment, we use the Arrhythmic data with original labels, not with preprocessed labels based on the ODDS repository. We extract normal class data as the training data and other data as the test data. After training the VAE, we calculate the anomaly score for each test data and also calculate the contribution degree of each dimension for the test data detected as an anomaly. For all the classes in which over 5 test data are detected as anomalies, we examined if the estimated contributing dimensions are relevant to the class to which the test data belongs. A t-distributed stochastic neighbor embedding (t-SNE) plot~\cite{maa2008} of the contribution degree is depicted in Fig.~\ref{fig:tsne} for classes 2, 3, 9, and 10 in which over 5 anomaly test data are detected. This indicates that the computed contribution degrees are associated with the class to which the test data belongs, excluding class 2.

We further analyzed the relationship of the characteristics of each class and estimated contributing dimensions in detail. Table~\ref{tab:cha} shows the characteristics of the classes summarized in Healio\footnote[4]{https://www.healio.com/cardiology/learn-the-heart}, which is a medical publishing website launched by The Wyanoke Group, and the estimated contributing dimensions according to the median of the contribution degree over the detected test data for extracting the common trend of the contribution degrees among the test data. In all the classes other than class 2, we found that the estimated contributing dimensions include dimensions relevant to the characteristic of the class.
As expected from the t-SNE plot, the estimated contributing dimensions varied in class 2 and observed common trend was trivial.
However, the fact also matches the characteristic of the class in which the symptoms are quite different from person to person. In class 3, although ST segment elevation, which should appear as an increase of SwaveAmp and TwaveAmp, was not observed in the top 5 contributing dimensions, QwaveAmp decreasing was observed in V3 and V4, which is indicative of myocardial infraction. In class 9, the increase of QRS duration and consecutive increase of Rwave width in V5 and V6 are observed. In class 10, the increase of RPwaveAmp and RPwave, which are the amplitude and the width of R' wave, respectively, should indicate the symptom of \textit{rsR' ``bunny ear'' pattern} described in the characteristic.

Note that contributing dimensions are estimated in a semi-supervised fashion with only data labeled as normal. This suggests that our algorithm enhances the VAEs for not only detecting anomalies accurately but also giving interpretations for identifying what type anomalies they are without labels of types.

\section{Conclusion}
For accurate and interpretable anomaly detection, we proposed a novel algorithm for estimating the contributing dimensions to the detected anomalies in deep learning-based methods. We adopted the variational autoencoder (VAE) for an anomaly detection algorithm, and for the detected anomalous data, our algorithm approximately explores a latent distribution of a normal data, which expresses how the anomalous data should appear if it were normal, and estimated the contributing dimensions and its degree in accordance with the log-likelihood computed by the obtained latent distribution. The estimated contributing dimensions should be exploited for identifying the cause of the detected anomalies. For example, even though labeled data in an anomalous situation cannot be easily obtained, clustering the anomalous data in accordance with the contribution degrees and using less labeled data may enable the causes of the detected anomalies to be automatically classified. Such approaches for automatically identifying the causes of the anomalies should be investigated in future work.

\bibliographystyle{aaai}
\bibliography{nwai}

\end{document}


\section{Appendix I. The algorithm proposed by~\cite{ike2017}.}
Their proposed algorithm for estimating contribution degrees to detected anomalies by the AE is based on the following assumption.

\noindent \textbf{Assumption.} The dimensions contributing to an anomaly are a part of the entire dimensions, and if the values of the contributing dimensions are fixed to plausible values in accordance with the learned relationships from the training data, the anomaly score output by the AE will become low.

On the basis of this assumption, they introduced vector $\bm{\eta}$ to express how much abnormal data $\bm{\bar{x}}$ differs from the plausible value for each dimension and define it as \textit{contribution degree}, which expresses how much the value of the dimension contributes to the detected anomaly. Consequently, the proposed algorithm for determining contribution degree $\bm{\eta}$ is expressed as the following minimization problem;
\begin{equation}\label{eq:spa}
\min_{\bm{\eta}} MSE(\bar{\bm{x}} - \bm{\eta}) + \lambda ||\bm{\eta}||_{1}.
\end{equation}
For determining the contributing dimensions, we set the threshold of contribution degree to $10^{-1}$ since sparse optimization sometimes fail to eliminate small noises in irrelevant dimensions.

\section{Appendix II. Comparison on accuracies for anomaly detection}

We examine the detection accuracy of the deep learning-based anomaly detection compared with conventional PCA-based anomaly detection in Table~\ref{tab:acc_ano}. For PCA, we define the squared length of the data projected to the abnormal subspace as the anomaly degree similar to the algorithm by~\cite{xu2009}. The number of the principal components is set to the same as the size of the latent spaces of the AE and the VAE. In all the algorithms, the threshold value is set to $\mu_{train} + 3 \sigma_{train}$, where $\mu_{train}$ and $\sigma_{train}$ are the mean and standard deviation values of the anomaly degrees calculated with each trained anomaly detection model and the training data. The experiment is executed similarly to~\cite{zha2016}. We randomly divide whole data by 1:1 and chose one as the training data by eliminating labeled data as anomaly and the other as the test data. We evaluate mean precision, recall, and F1 score through 20 runs. As shown in Table~\ref{tab:acc_ano}, the deep learning-based algorithms tend to be more accurate than the PCA. Although more careful tuning of parameters will further improve the accuracies of deep learning-based models, we evaluated with simple parameter settings under the condition that the sizes of latent space are aligned with PCA.

\begin{table}[t]
\caption{Detection accuracies.}
\scalebox{0.85}{ 
\begin{tabular}{c|c|c|c|c|c}
\toprule[0.5mm]
Method & Metric & Arrhythmia & MNIST & Musk & NSL-KDD \\
\midrule
 & Precision & 0.27 & \textbf{0.73} & 0.65 & 0.97  \\
PCA & Recall & \textbf{0.78} & 0.17 & \textbf{1} & 0.83 \\
 & $F_1$ & 0.40 & 0.27 & 0.79 & 0.89 \\
\midrule
 & Precision & \textbf{0.58} & 0.72 & \textbf{0.82} & \textbf{0.98}  \\
AE & Recall & 0.32 & 0.17 & \textbf{1} & 0.85 \\
 & $F_1$ & 0.41 & 0.28 & \textbf{0.90} & 0.91 \\
\midrule
 & Precision & 0.39 & 0.68 & 0.77 & 0.96 \\
VAE & Recall & 0.47 & \textbf{0.23} & \textbf{1} & \textbf{0.89} \\
 & $F_1$ & \textbf{0.43} & \textbf{0.34} & 0.87 & \textbf{0.93} \\
\bottomrule[0.5mm]
\end{tabular}
}
\label{tab:acc_ano}
\end{table}

\bibliographystyle{aaai}
\bibliography{nwai}